\documentclass{llncs}

\usepackage{eccv}

\usepackage{eccvabbrv}

\usepackage{graphicx}
\usepackage{booktabs}

\usepackage[accsupp]{axessibility}  

\usepackage{hyperref}

\usepackage{orcidlink}

\usepackage{amsmath,amsfonts}
\usepackage{algorithmic}
\usepackage{algorithm}
\usepackage{array}
\usepackage{textcomp}
\usepackage{url}
\usepackage{verbatim}
\usepackage{multirow}
\usepackage{makecell}
\usepackage{caption}
\usepackage{cite}
\usepackage{xcolor}
\usepackage{comment}

\usepackage[leftcaption]{sidecap}
\sidecaptionvpos{figure}{c}

\begin{document}
\renewcommand{\footnoterule}{%
  \vspace*{-\footnotesep}%
  \hrule width 0.3\textwidth height 0.4pt %
  \vspace*{2pt}%
}

\title{Road Maps as Free Geometric Priors: \\ Weather-Invariant Drone Geo-Localization \\ with GeoFuse}


\author{
Yunsong Fang\inst{1} \and
Tingyu Wang\inst{2} \and
Zhedong Zheng\inst{1}
}

\authorrunning{Y.~Fang et al.}

\institute{
University of Macau, Macau SAR, China\\
\email{\{mc45296, zhedongzheng\}@um.edu.mo}
\and
Hangzhou Dianzi University, Hangzhou, China\\
\email{tingyu.wang@hdu.edu.cn}
}

\maketitle

\begin{abstract}
  Drone-view geo-localization aims to match a query drone image, often captured under adverse weather conditions (\eg, rain, snow, fog), against a gallery of geo-tagged satellite images. Weather-induced degradations in the drone view, such as noise, reduced visibility, and partial occlusions, severely exacerbate the intrinsic cross-view domain gap. While prior methods predominantly rely on weather-specific architectures or data augmentations, they have largely overlooked road map data, a readily available modality that provides strong, inherently weather-invariant geometric layout cues (\eg, road networks and building footprints) at negligible additional cost. We introduce GeoFuse, a cross-modal fusion framework that integrates precisely aligned road map tiles with satellite imagery to yield more discriminative and weather-resilient representations. We first augment the existing University-1652 and DenseUAV benchmarks with geo-aligned road maps, supplying structural priors robust to meteorological variations. Building on this, we propose a flexible fusion module that combines satellite and road map features via token-level and channel-level interactions, with a lightweight dynamic gating mechanism that adaptively weights modality contributions per instance. Finally, we employ class-level cross-view contrastive learning to promote robust alignment between weather-degraded drone features and the fused satellite-roadmap representations. Extensive experiments under diverse weather conditions show that GeoFuse consistently outperforms state-of-the-art methods, achieving $+3.46\%$ and $+23.18\%$ Recall@1 accuracy on the University-1652 and DenseUAV benchmarks, respectively. Our code will be released at \url{https://github.com/YsongF/GeoFuse}.
\end{abstract}

\section{Introduction}
\label{sec:intro}

Drone-based geo-localization intends to determine the precise geographic location of a drone-captured image by matching it against reference geo-tagged satellite images~\cite{zheng2020university, wang2021each, dai2023vision, wang2024learning, wen2025WeatherPrompt}. This task plays a critical role in numerous real-world applications, including autonomous navigation, urban monitoring, emergency response, and environmental mapping~\cite{durgam2024cross, xia2025cross}. Compared with traditional Global Navigation Satellite Systems (GNSS)-based localization, visual geo-localization provides an alternative that remains effective even when GPS signals are weak or unavailable, such as in dense urban areas or remote regions~\cite{zhu2021vigor, sun2025cgsi, chen2025multi}. However, due to significant viewpoint differences, appearance variations, and environmental changes between drone and satellite images, achieving reliable matching across domains remains highly challenging. Recent deep learning-based approaches~\cite{deuser2023sample4geo, xia2024enhancing, xie2025self} have achieved remarkable progress by learning shared representations between the drone and satellite views. However, real-world multiple weather conditions, such as fog, rain, snow, and illumination changes, introduce visual noise, visibility loss, and scene occlusions in drone imagery, thereby aggravating feature degradation and widening the cross-view domain gap. Existing methods~\cite{wang2024multiple, zhou2025cdm} mainly rely on RGB imagery from drone and satellite views, and are typically trained on clear-weather datasets, which restricts their generalization across varying weather and environmental conditions. 

\begin{SCfigure}[0.6][!t]
  {\vspace{-.2in}\includegraphics[width=0.5\linewidth]{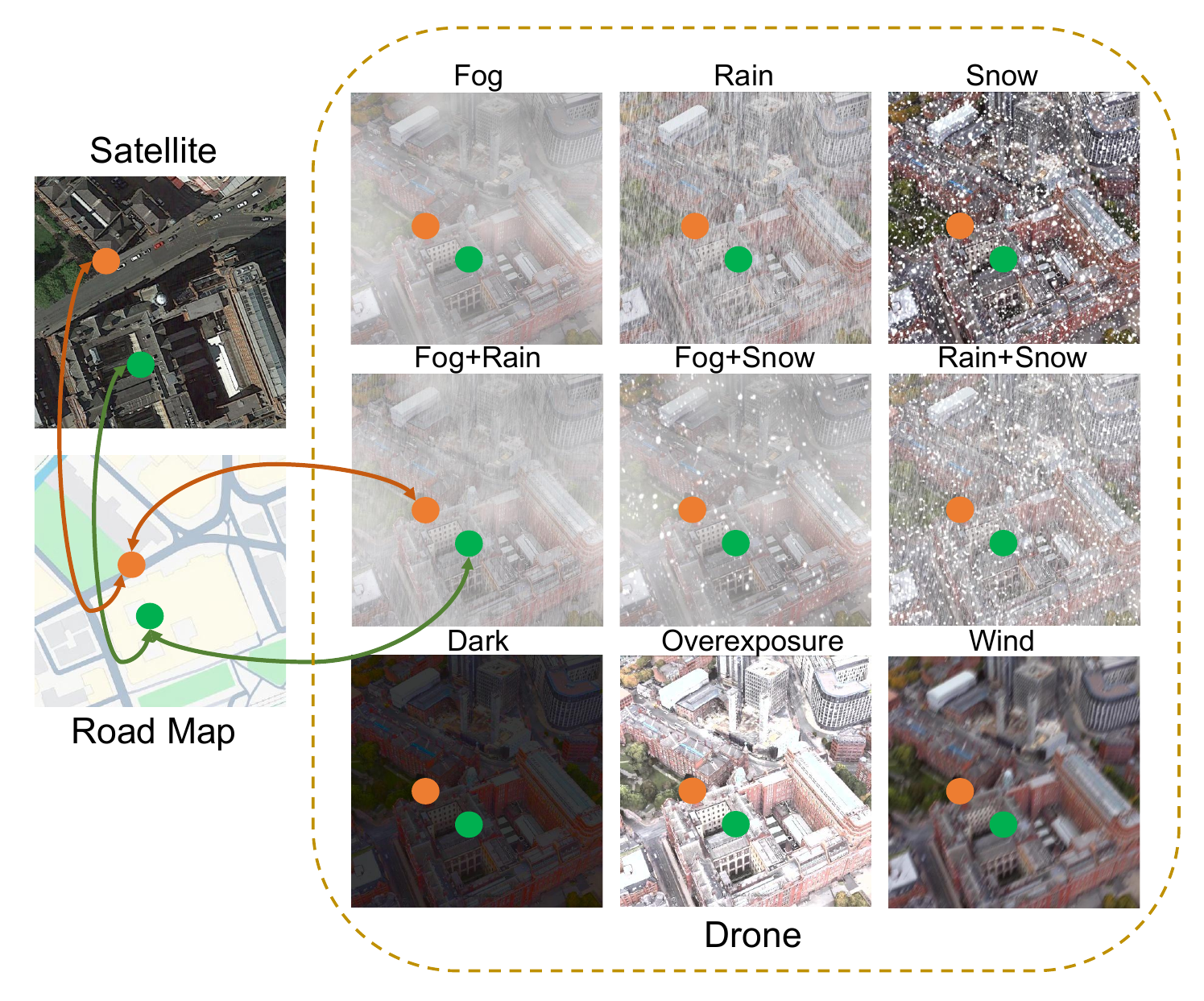}}
  \caption{Intuitive illustration of how the ``free'' road map provides geometric structural priors to assist robust matching between drone-view and satellite-view images, particularly under adverse weather conditions.}
  \label{fig_1}
  \vspace{-.15in}
\end{SCfigure}

In an attempt to mitigate the weather impact, recent studies have mainly focused on architectures tailored for specific weather conditions or on extensive data augmentation. In addition, researchers have resorted to cross-modal fusion strategies, incorporating complementary modalities (\eg, video, text) to enhance discrimination ability. For example, Ju \etal~\cite{ju2025video2bev} aggregate multi-frame features to obtain more consistent and weather-robust cross-platform representations by leveraging drone videos. Wen \etal~\cite{wen2025WeatherPrompt} employ weather and drone-view related textual descriptions as auxiliary input, enabling the model to adapt visual representations according to different weather scenarios. However, existing approaches have largely neglected road map data, a readily accessible modality that provides weather-invariant spatial geometric cues, which can complement satellite imagery at minimal cost, as shown in~\cref{fig_1}. 

To address these challenges, we introduce GeoFuse, a novel cross-modal fusion framework that leverages road map semantics alongside satellite imagery. First, we enrich satellite representations with spatial geometric cues that remain robust under varying weather conditions by incorporating geo-aligned road map tiles. In particular, we extend popular benchmarks, \ie, University-1652 and DenseUAV, with corresponding road map data, providing structural priors that complement visual features and enhance cross-platform matching. Central to GeoFuse is a flexible fusion module that integrates satellite and road map features through interactions at both token and channel levels, controlled by a lightweight dynamic gating mechanism that adaptively modulates the influence of each modality for individual instances. Furthermore, to ensure robust alignment across modalities even in degraded conditions, we employ class-level cross-view contrastive learning, encouraging drone features to align consistently with the fused satellite-roadmap representations. These components enable the network to produce more discriminative and robust representations across varying weather conditions. Experiments validate that GeoFuse consistently surpasses current state-of-the-art methods, achieving approximately $+3.46\%$ and $+23.18\%$ Recall@1 gains on the University-1652 and DenseUAV benchmarks against diverse weather conditions, respectively. 
In summary, our primary contributions are as follows:

$\bullet$ Leveraging freely available road maps alongside drone and satellite images, we propose GeoFuse, an adaptive cross-modal fusion framework that performs dual-level feature fusion at both token and channel scales through a lightweight gating design. Furthermore, a class-level cross-modal contrastive objective is designed to reinforce geometric alignment consistency and enhance the discriminability of cross-view representations.

$\bullet$ To validate the method, we extend the existing benchmarks, \ie, University-1652 and DenseUAV, by introducing geo-aligned road maps paired with satellite imagery, providing weather-invariant geometric cues that enrich cross-platform representations and improve the reliability of cross-view geo-localization.

$\bullet$ Extensive experiments on both non-dense (University-1652) and dense (DenseUAV) datasets validate that GeoFuse achieves state-of-the-art recall accuracy and average precision, exhibiting strong generalization and robustness across diverse weather conditions. 

\section{Related Work}

Cross-view geo-localization aims to infer the geographic location of ground-level or aerial-view images by matching them with geo-referenced satellite imagery from the same region but under drastically different viewpoints~\cite{zheng2020university, wen2025WeatherPrompt}. Early methods rely on hand-crafted features to extract and match distinctive structures (\eg, buildings), followed by geometric verification~\cite{tian2017cross, wang2023fine}. However, these approaches struggle with viewpoint, illumination, and texture variations~\cite{cai2018feature}. The advent of deep learning has shifted the paradigm to CNN-based methods, which learn domain-invariant representations via joint embedding~\cite{deuser2023sample4geo}. Workman \etal~\cite{workman2015wide} pioneer a Siamese CNN for view alignment, introducing CVUSA. Liu and Li~\cite{liu2019lending} extend this with directional cues and propose CVACT. Both datasets focus on paired ground-satellite images for supervised matching. Zheng \etal~\cite{zheng2020university} advance the field with University-1652, the first benchmark incorporating drone-view images alongside ground and satellite views, enabling multi-platform cross-view geo-localization. Subsequent CNN advancements (ResNet~\cite{he2016deep}, DenseNet~\cite{huang2017densely}, ConvNeXt~\cite{liu2022convnet}) improve feature extraction, while attention mechanisms~\cite{shi2019spatial, gao2025semantic} and Transformer-based models enhance spatial alignment and global context. Pioneering Transformer works include Yang \etal~\cite{yang2021cross} with position-aware encoding, TransGeo~\cite{zhu2022transgeo} with attention-guided cropping, FSRA~\cite{dai2021transformer} for region segmentation and alignment, TransFG~\cite{zhao2024transfg} with gradient-guided aggregation, and Swin Transformer variants~\cite{liang2025dstg} for efficient embeddings. Dai \etal~\cite{dai2023vision} introduce DenseUAV, a densely sampled UAV-satellite dataset addressing sparsity in prior benchmarks, using Transformer-based metric learning for improved discrimination. Recent efforts focus on multi-weather pretraining, domain adaptation, and augmentation for robustness under adverse conditions (fog, rain, illumination)~\cite{liu2025novel, zhao2025p2fcn, wu2025ccigeo}. Contrastive learning dominates current approaches, with objectives like triplet loss, InfoNCE, and instance loss optimizing embeddings by pulling positives closer and pushing negatives apart~\cite{ahn2024bridging, xia2024enhancing}. However, sample-level contrasts often neglect higher-level semantics and remain sensitive to noisy negatives or imbalanced correspondences, especially in challenging weather. To mitigate this, we propose a class-level contrastive loss that leverages category associations for cross-view alignment.

Recently, multi-modality learning provides robust and effective solutions for cross-view geo-localization, particularly under challenging weather conditions. Deuser \etal~\cite{deuser2023orientation} incorporate orientation as an auxiliary modality, using satellite views to estimate drone-view directions as pseudo-labels for contrastive learning, enhancing cross-view consistency. Other works target fine-grained localization via 3-DoF camera pose estimation~\cite{shi2022accurate, wang2023fine}, or leverage video modalities for richer temporal cues~\cite{vyas2022gama, zhang2023cross, ju2025video2bev}. For instance, Vyas \etal~\cite{vyas2022gama} introduce the GAMa dataset with ground videos paired to satellite images and propose hierarchical clip-to-video matching. Zhang \etal~\cite{zhang2023cross} aggregate temporal features with attention and sequential dropout for variable-length sequences. Video2BEV~\cite{ju2025video2bev} transforms drone videos into Bird’s Eye Views via Gaussian Splatting and diffusion-based hard negative generation, validated on the new UniV dataset extending University-1652. With large multimodal models like CLIP~\cite{radford2021learning}, recent methods~\cite{chu2024towards, sun2025cgsi, ye2025cross} integrate textual semantics to guide feature alignment, leveraging image-text contrast for improved discriminability and robustness. However, existing approaches largely overlook accessible road map data, which provides stable geometric priors invariant to weather, illumination, or viewpoint changes. In this work, we extend vision-language models by incorporating road maps as geometric guidance for satellite imagery and propose a novel road-satellite fusion strategy to enhance feature correspondence and cross-view matching robustness.

\section{Proposed Method}

\begin{figure}[!t]
\vspace{-.15in}
\centering
\includegraphics[width=0.95\textwidth]{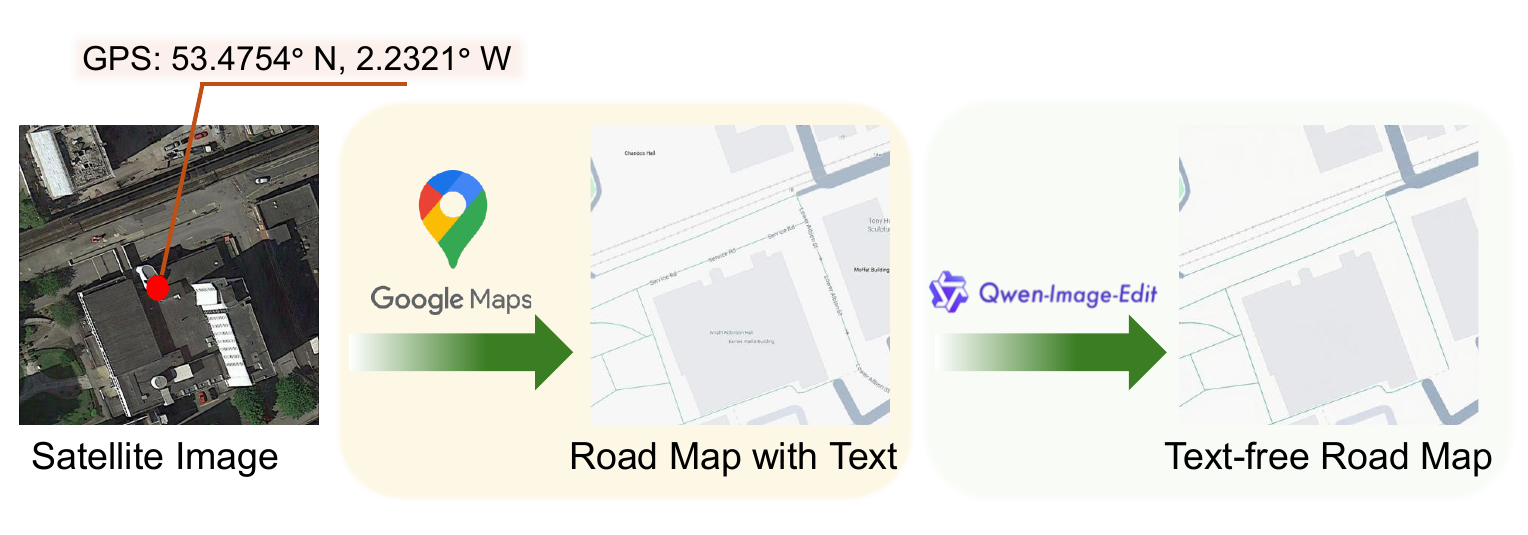}
\vspace{-.25in}
\caption{Road map collection process. The obtained text-free road maps preserve the underlying road topology and spatial layout while eliminating semantic cues from map annotations.}
\vspace{-.2in}
\label{fig_2}
\end{figure}

\subsection{Road Map Curation}
\label{sec:road}
To acquire the required road maps, we first extract the geographic coordinates of satellite images from public benchmarks for drone-view geo-localization, \ie, University-1652~\cite{zheng2020university} and DenseUAV~\cite{dai2023vision}. Based on these coordinates, corresponding road map tiles are retrieved via the Google Maps API\footnote{\tiny\url{https://www.google.com/maps}} at an optimally selected zoom level that balances the visibility of the road network structure and the spatial coverage consistent with the satellite images used in our experiments. As illustrated in~\cref{fig_2}, the collected satellite image coordinates are first used to query the Google Maps service to obtain the corresponding road maps, which initially contain textual information such as street names and geographical annotations. However, such textual elements introduce unintended cues that could bias the geo-localization model. To address this issue, we employ the Qwen-Image-Edit~\cite{wu2025qwen} model to remove all textual content from the retrieved road maps. 
Finally, the resulting text-free road maps are geographically aligned with the corresponding satellite images as weather-invariant geometric layout priors.

\noindent\textbf{Discussion. Why remove textual content?} The removal of textual content from road maps serves multiple critical purposes in enhancing the robustness of our geo-localization framework. Primarily, it prevents information leakage, as textual elements like street names or labels could inadvertently provide explicit identifiers that models might exploit as shortcuts during training, leading to overfitting on dataset-specific artifacts rather than learning generalizable visual features. This is particularly problematic in benchmarks like University-1652 and DenseUAV, where such annotations might correlate with known locations, potentially inflating performance metrics without true cross-view understanding. Additionally, by eliminating text, the maps become more agnostic to specific locales, facilitating application to unknown or novel environments where textual information may be absent, inconsistent, or unavailable. This design choice thereby strengthens the model's generalization ability, ensuring that it relies solely on structural and geometric cues, such as road intersections and layouts, for inference, which is essential for real-world deployment in diverse, uncharted geographic settings.

\begin{figure}[!t]
\vspace{-.2in}
\centering
\includegraphics[width=0.95\textwidth]{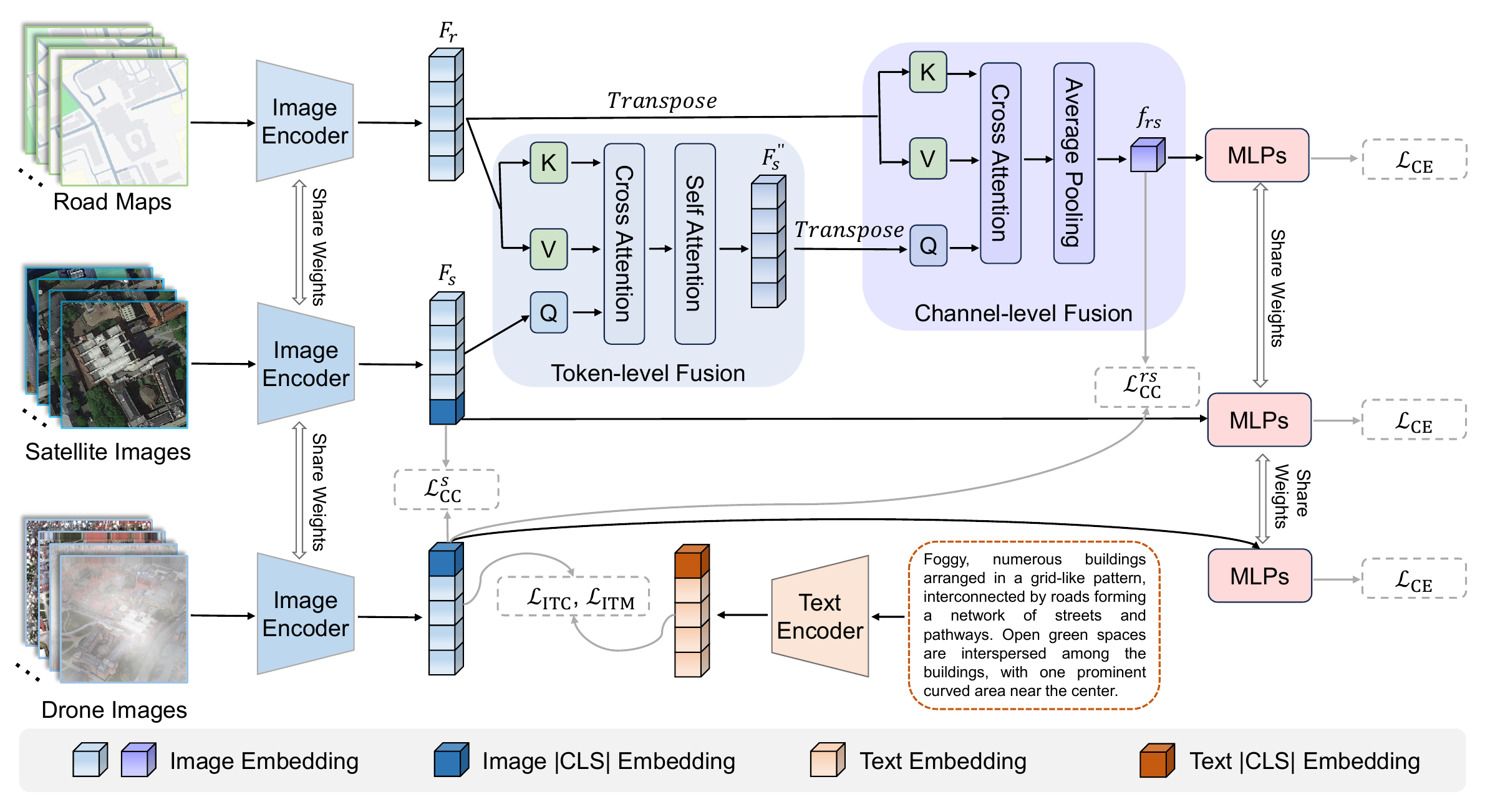}
\vspace{-.15in}
\caption{The brief overview of GeoFuse. Given the visual input triplet of road maps, satellite images, and drone images, together with multi-weather textual descriptions of the drone images, we first extract visual and textual features through a shared-weight image encoder and a dedicated text encoder. The framework then processes two branches: a drone-text alignment branch for multi-weather cross-modal matching, and a satellite-road fusion branch for geometric-aware representation learning. The satellite-road fusion branch fuses the satellite feature $F_s$ and the road feature $F_r$ as $f_{rs}$. Finally, we apply the alignment losses to supervise these representations.}
\vspace{-.2in}
\label{fig_3}
\end{figure}

\subsection{GeoFuse}
We introduce a cross-modal fusion framework to harness road map guidance for multi-weather drone-view geo-localization, as shown in~\cref{fig_3}. 
The framework consists of a cross-modal learning module and a joint optimization module. The cross-modal learning module includes an image encoder and a text encoder for modality-specific feature extraction, and is composed of two branches: one aligns multi-weather drone images with their corresponding textual descriptions, while the other integrates satellite imagery with geographically aligned road maps to achieve unified multi-view representation learning. The multimodal feature is then passed through a lightweight classification head that shares weights. To guide the direction of cross-modal learning, a comprehensive loss module is designed, integrating cross-modality matching loss, contrastive loss, and cross-entropy loss. This module jointly constrains feature learning from both inter-modal association and intra-modal discrimination. These designs enable the framework to enhance the accuracy and robustness of drone-view geo-localization in adverse weather conditions.

The cross-modal learning module aims to model correspondence and integrate complementary semantics from multiple modalities to enhance the robustness and discriminability of feature representations. It consists of two branches: a drone-text alignment branch and a satellite-road fusion branch. The drone-text alignment branch adopts a dual-encoder architecture consisting of an image encoder and a text encoder, which follows the existing work~\cite{wen2025WeatherPrompt}. The image encoder extracts visual representations from multi-weather drone images, while the text encoder encodes the corresponding textual descriptions into text embeddings. Both representations are projected into a shared embedding space to capture semantic correspondences across modalities. In the satellite-road fusion branch, both satellite-view features and road map features are extracted with the same image encoder with shared weights. We fuse these two types of features via an attention-based fusion mechanism that consists of two complementary components: token-level fusion for modeling spatial correspondences and channel-level fusion for capturing interdependency across feature channels. This mechanism is equipped with a lightweight dynamic gating scheme to enhance flexibility, enabling adaptive modality weighting per input. The cross-modal features from both branches are then fed into a single-layer MLP classifier to generate unified embeddings and predict the geographic location categories.

\noindent\textbf{Token-level Fusion.}
Let $F_s \in \mathbb{R}^{N \times D}$ and $F_r \in \mathbb{R}^{N \times D}$ denote the normalized image embeddings of satellite-view and roadmap images extracted from the image encoder, where $N$ is the number of tokens per image, and $D$ is the feature dimension. To capture fine-grained spatial correspondences between the two modalities, we perform cross-attention followed by self-attention refinement. Since the road map serves as an auxiliary modality providing geometric layout cues for the satellite image, it is employed as the key and value in the cross-attention to update the satellite image tokens:
\vspace{-4pt}
\begin{equation}
F'_s = F_s + w_1 \cdot \text{MHA}(F_s, F_r, F_r),
\label{eq:token_cross}
\end{equation}
where $\text{MHA}(\cdot)$ denotes multi-head attention function with $Q,K,V$ inputs, and $w_1$ is a learnable gating parameter that modulates the contribution of the road map in the fusion process. Subsequently, a self-attention layer refines the fused tokens:
\vspace{-4pt}
\begin{equation}
F''_s = F'_s + w_2 \cdot \text{MHA}(F'_s, F'_s, F'_s),
\label{eq:token_self}
\end{equation}
where $w_2$ is another learnable gating scalar, and $F''_s \in \mathbb{R}^{N \times D}$ denotes the output of the token-level fusion stage. Each attention block follows the standard Transformer~\cite{vaswani2017attention} design, which includes feed-forward and normalization layers. 

\noindent\textbf{Channel-level Fusion.}
On top of the token-level fusion, we leverage road map feature channels to guide channel-level fusion, enabling the model to capture cross-modal interactions across feature dimensions. We first transpose the token-level fusion output $F''_s$ and the road map embeddings $F_r$ into ${F''_s}^\top \in \mathbb{R}^{D \times N}$ and $F_r^\top \in \mathbb{R}^{D \times N}$, respectively, treating feature channels as the sequence dimension for the subsequent fusion. The fused features are then updated via channel-level cross-attention:
\vspace{-4pt}
\begin{equation}
F_{rs} = {F''_s}^\top + w_3 \cdot \text{MHA}({F''_s}^\top, F_r^\top, F_r^\top)  \quad \in \mathbb{R}^{D \times N},
\end{equation}
where ${F''_s}^\top$ serve as queries, $F_r^\top$ serve as keys and values, and $w_3$ is a learnable gating parameter controlling the contribution of cross-channel interaction.

Finally, the dual-level fused feature is obtained by applying global average pooling over the token dimension:
\begin{equation}
f_{rs} = avgpool(F_{rs}) \quad \in \mathbb{R}^{D},
\end{equation}
which is subsequently fed into a lightweight MLP-based classifier for downstream optimization.

\subsection{Optimization Objectives}
The joint training objectives of our multi-modal interaction framework consist of three components. First, for the multi-weather drone-text branch, we adopt two cross-modal objectives following Wen \etal~\cite{wen2025WeatherPrompt}, \ie, a contrastive loss $\mathcal{L}_\text{ITC}$ and a matching loss $\mathcal{L}_\text{ITM}$, which help align visual and textual embeddings. We denote their sum as $\mathcal{L}_\text{IT} = \mathcal{L}_\text{ITC} + \mathcal{L}_\text{ITM}$. Second, to enhance cross-view feature discriminability and robustness under challenging conditions, we introduce class-level contrastive loss $\mathcal{L}_\text{CC}$ based on the InfoNCE~\cite{oord2018representation}, which leverages class-level anchors for more stable feature alignment. Finally, the instance loss $\mathcal{L}_\text{CE}$, which is a softmax cross-entropy  loss with weight-shared classifiers~\cite{zheng2020university}, is applied to supervise the drone, satellite, and fused multi-modal embeddings.

\noindent\textbf{Class-level Contrastive Loss.} In previous drone-based geo-localization methods, contrastive learning is performed only among samples within the same batch, which makes the training effectiveness highly sensitive to batch size. To overcome this limitation, we adopt a class-level contrastive learning strategy, where each category is represented by a pre-computed anchor feature that provides a stable reference for aligning features across different views. Since satellite and road map imagery for the same geographic location remain relatively stable compared to drone views, we precompute two types of category-level anchors before training.  In particular, let $\mathcal{Y} = \{1,2,\dots,C\}$ denote the $C$ predefined location classes in the training set, where $y_c$ represents the label of the $c$-th location. 
For the $c$-th location, let ${f}_s^c$ and ${f}_{rs}^c$ denote the feature extracted from the satellite image and its corresponding satellite-road fused representation, respectively. If there is typically one satellite-road pair per location, these represent the individual features; otherwise, for locations with more than one image, they denote the mean features. 
The satellite feature ${f}_s^c \in \mathbb{R}^{D}$ is obtained from the [CLS] embedding of the image encoder output, followed by normalization and a lightweight classification head from which the intermediate feature is extracted. 
The anchor set is then established in a dictionary form, corresponding to the satellite and satellite-road fused modalities:
\vspace{-4pt}
\begin{equation}
\mathcal{A} = \left\{ (y_c, {f}_s^c, {f}_{rs}^c) \right\}_{c=1}^{C},
\end{equation}
where the anchor set stores two representative features per class.

For a training batch of drone images with batch size $B$, let ${f}_d^i \in \mathbb{R}^{D}$ denote the feature of the $i$-th drone sample ($i = 1,2,\dots,B$), which is extracted via the same pipeline as the satellite feature ${f}_s^c$. Let $y_d^i \in \mathcal{Y}$ denote the corresponding geographic location class label. To align drone features with two pre-defined class-level anchors, we design a class-level contrastive loss that enforces intra-class similarity and inter-class discrimination, alleviating batch size sensitivity. Two similarity, \ie, $\mathbf{S}^s, \mathbf{S}^{rs} \in \mathbb{R}^{B \times C}$, are computed between the drone features $\{{f}_d^i\}_{i=1}^B$ and the anchor features, $\{{f}_s^c\}_{c=1}^C$ and $\{{f}_{rs}^c\}_{c=1}^C$, respectively. Each element, $S_s^{i,c}$ and $S_{rs}^{i,c}$, represents the similarity between the $i$-th drone feature and the $c$-th anchor feature in the satellite and fused modalities, respectively:
\vspace{-4pt}
\begin{equation}
S_s^{i,c} = \frac{{f}_d^i \cdot {f}_s^{c\top}}{\tau}, \quad
S_{rs}^{i,c} = \frac{{f}_d^i \cdot {f}_{rs}^{c\top}}{\tau}.
\end{equation}
where $\tau = 0.07$ is a temperature hyper-parameter. We then construct a positive mask $M \in \{0,1\}^{B \times C}$ to identify matching class pairs, defined as:
\vspace{-4pt}
\begin{equation}
M_{i,c} =
\begin{cases}
1, & \text{if } y_d^i = y_c, \\
0, & \text{otherwise}.
\end{cases}
\end{equation}

The class-level contrastive losses for satellite and fused modalities within one batch are first computed separately:
\vspace{-4pt}
\begin{equation}
\begin{aligned}
\mathcal{L}_{\text{CC}}^s &= -\frac{1}{B} \sum_{i=1}^B \log\left( \frac{\sum_{c=1}^C \exp(S_s^{i,c}) \cdot M_{i,c}}{\sum_{c=1}^C \exp(S_s^{i,c})} \right), \\
\mathcal{L}_{\text{CC}}^{rs} &= -\frac{1}{B} \sum_{i=1}^B \log\left( \frac{\sum_{c=1}^C \exp(S_{rs}^{i,c}) \cdot M_{i,c}}{\sum_{c=1}^C \exp(S_{rs}^{i,c})} \right).
\end{aligned}
\end{equation}
To prevent division by zero during implementation, both the numerator and denominator in the above two loss functions are clamped to a minimum value of $10^{-8}$. The final class-level contrastive loss $\mathcal{L}_{\text{CC}}$ is the sum of the two modality-specific losses:
\vspace{-4pt}
\begin{equation}
\mathcal{L}_{\text{CC}} = \mathcal{L}_{\text{CC}}^s + \mathcal{L}_{\text{CC}}^{rs},
\end{equation}
which encourages each drone feature to be pulled closer to the anchor of its own class while being pushed away from anchors of other classes, providing a stable supervision signal independent of the batch composition.

\noindent\textbf{Final Objective.}  The total loss is formulated as:
\vspace{-4pt}
\begin{equation}
\mathcal{L}_{\text{total}} = \mathcal{L}_{\text{IT}} + \mathcal{L}_{\text{CE}} + \lambda \cdot \mathcal{L}_{\text{CC}},
\end{equation}
where $\lambda = 0.10$ is a hyper-parameter controlling the relative contribution of the class-level contrastive loss $\mathcal{L}_{\text{CC}}$. This multi-modal joint objective enables the model to learn weather-invariant representations guided by geometric cues, thereby enhancing generalization and localization robustness across diverse weather conditions.

\section{Experiment}
\subsection{Dataset and Implementation Details}
\noindent\textbf{Dataset.} We evaluate the proposed model on both non-dense and dense drone-based geo-localization datasets according to the spatial sampling density. 
For non-dense datasets, target locations are spatially independent and visually distinctive. We select the widely-adopted University-1652 dataset~\cite{zheng2020university}, containing drone-view, satellite-view, and street-view images of 1,652 buildings from 72 universities worldwide. It includes 701 training buildings and 951 testing buildings, with 1 satellite image and 54 drone images per building. In contrast, the dense dataset explicitly includes target locations with overlapping, where neighboring viewpoints often appear highly similar yet correspond to different locations, increasing the challenge of discriminative feature learning and cross-view matching. We evaluate methods on DenseUAV~\cite{dai2023vision}, a large-scale benchmark for UAV self-positioning in low-altitude urban environments. Collected from 14 university campuses in Zhejiang, China, it comprises 27,297 multi-height drone-view and multi-temporal satellite images, split into 2,256 training classes and 777 strictly disjoint test classes. We report Recall@1 (R@1) and Average Precision (AP). R@1 measures the proportion of correctly localized images in the top-1 retrieval result, while AP is the area under the Precision-Recall curve. Higher values indicate better performance. 

\noindent\textbf{Implementation Details.} We build on WeatherPrompt~\cite{wen2025WeatherPrompt} and adopt X-VLM~\cite{zeng2021multi} (pre-trained on 4M data with COCO~\cite{lin2014microsoft} and VG~\cite{krishna2017visual} annotations) as the backbone. It employs Swin Transformer~\cite{liu2021swin} as the image encoder and BERT~\cite{devlin2019bert} as the text encoder. Images are resized to 384$\times$384 pixels and divided into 32$\times$32 non-overlapping patches. The model is trained for 210 epochs with batch size 8 using SGD (momentum 0.9, weight decay 0.0005). The learning rate is decayed by 0.1 at epoch 120 and by 0.01 at epoch 180. Training augmentations include random cropping and horizontal flipping for both views, plus weather transformations for drone-view images. Satellite-drone pairs undergo synchronized horizontal flipping and rotation before feature fusion. During testing, similarity is computed via Euclidean distance. All experiments are implemented in PyTorch~\cite{paszke2019pytorch} on a single NVIDIA A100 GPU.

\begin{table}[!t]
\caption{Performance (R@1 ($\%$) and AP ($\%$)) on University-1652~\cite{zheng2020university} for Drone \ensuremath{\rightarrow} Satellite and Satellite \ensuremath{\rightarrow} Drone tasks under multiple weathers. Best results are highlighted in bold. Methods marked with $^*$ adopt the model weights publicly available in their official GitHub repositories. \label{tab:table1}}
\vspace{-10pt}
\centering
\small
\renewcommand{\arraystretch}{1.0}  
\setlength{\tabcolsep}{1.2pt}
\resizebox{\linewidth}{!}{
\begin{tabular}{l|l|c c|c c|c c|c c|c c|c c|c c|c c|c c|c c|c c}
\toprule
\multirow{2}{*}{Method} & \multirow{2}{*}{Backbone} & \multicolumn{2}{c|}{Normal} & \multicolumn{2}{c|}{Fog} & \multicolumn{2}{c|}{Rain} & \multicolumn{2}{c|}{Snow} & \multicolumn{2}{c|}{Fog+Rain} & \multicolumn{2}{c|}{Fog+Snow} & \multicolumn{2}{c|}{Rain+Snow} & \multicolumn{2}{c|}{Dark} & \multicolumn{2}{c|}{Over-exp} & \multicolumn{2}{c|}{Wind} & \multicolumn{2}{c}{Mean}\\
& & R@1 & AP & R@1 & AP & R@1 & AP & R@1 & AP & R@1 & AP & R@1 & AP & R@1 & AP & R@1 & AP & R@1 & AP & R@1 & AP & R@1 & AP\\
\midrule
\multicolumn{24}{c}{Drone \ensuremath{\rightarrow} Satellite}\\
\midrule
LRFR*~\cite{gan2025learning} & ConvNeXt & \textbf{94.13} & \textbf{95.09} & 48.68 & 53.54 & 34.38 & 38.80 & 31.34 & 36.18 & 7.64 & 10.44 & 4.52 & 6.68 & 31.48 & 35.85 & 10.45 & 13.29 & 65.49 & 69.59 & \textbf{83.74} & \textbf{86.15} & 41.19 & 44.56\\
Zheng \etal~\cite{zheng2020university} & ResNet-50 & 67.83 & 71.74 & 60.97 & 65.23 & 60.29 & 64.61 & 55.58 & 60.09 & 54.75 & 59.40 & 44.85 & 49.78 & 57.61 & 62.03 & 39.70 & 44.65 & 51.85 & 56.75 & 58.28 & 62.83 & 55.17 & 59.17\\
He \etal~\cite{he2016deep} & ResNet-101 & 70.07 & 73.04 & 63.87 & 68.22 & 63.34 & 67.59 & 59.75 & 64.15 & 57.45 & 62.12 & 48.31 & 53.28 & 60.25 & 64.68 & 46.12 & 51.02 & 56.34 & 61.23 & 62.13 & 66.63 & 58.76 & 63.29\\ 
Huang \etal~\cite{huang2017densely} & DenseNet121& 69.48 & 73.26 & 64.25 & 68.47 & 63.47 & 67.64 & 59.29 & 63.70 & 59.68 & 64.13 & 50.41 & 55.20 & 60.21 & 64.57 & 48.57 & 53.41 & 54.04 & 58.88 & 60.74 & 65.14 & 59.01 & 63.44\\
Safe-Net*~\cite{lin2024self} & Vit-S~\cite{dosovitskiy2020image} & 86.98 & 88.85 & 82.12 & 86.10 & 67.13 & 68.90 & 60.50 & 63.01 & 54.80 & 58.73 & 32.12 & 39.77 & 25.83 & 26.40 & 41.10 & 44.13 & 69.87 & 71.15 & 74.32 & 76.58 & 60.48 & 63.36\\
Liu \etal~\cite{liu2021swin} & Swin-T & 69.27 & 73.18 & 66.46 & 70.52 & 65.44 & 69.60 & 61.79 & 66.23 & 63.96 & 68.21 & 56.44 & 61.07 & 62.68 & 67.02 & 50.27 & 55.18 & 55.46 & 60.29 & 63.81 & 68.17 & 61.56 & 65.95\\
Pan \etal~\cite{pan2018two} & IBN-Net & 72.35 & 75.85 & 66.68 & 70.64 & 67.95 & 71.73 & 62.77 & 66.85 & 62.64 & 66.84 & 51.09 & 55.79 & 64.07 & 68.13 & 50.72 & 55.53 & 57.97 & 62.52 & 66.73 & 70.68 & 62.30 & 66.46\\
LPN~\cite{wang2021each} & ResNet-50 & 74.33 & 77.60 & 69.31 & 72.95 & 67.96 & 71.72 & 64.90 & 68.85 & 64.51 & 68.52 & 54.16 & 58.73 & 65.38 & 69.29 & 53.68 & 58.10 & 60.90 & 65.27 & 66.46 & 70.35 & 64.16 & 68.14\\
Muse-Net~\cite{wang2024multiple} & ResNet-50 $\times 2$ & 74.48 & 77.83 & 69.74 & 73.24 & 70.55 & 74.14 & 65.72 & 69.70 & 65.59 & 69.94 & 54.69 & 59.24 & 66.64 & 70.55 & 53.85 & 58.49 & 61.05 & 65.51 & 69.45 & 73.22 & 65.15 & 69.16\\
\hline
Baseline~\cite{wen2025WeatherPrompt} & X-VLM & 82.78 & 85.18 & 81.46 & 84.03 & 80.34 & 83.11 & 77.60 & 80.67 & 78.75 & 81.69 & 73.38 & 76.94 & 78.41 & 81.40 & 67.22 & 71.06 & 74.20 & 77.63 & 77.26 & 80.27 & 77.14 & 80.20\\
Ours & X-VLM & 85.26 & 87.44 & \textbf{84.72} & \textbf{86.96} & \textbf{83.61} & \textbf{86.02} & \textbf{81.38} & \textbf{84.08} & \textbf{82.26} & \textbf{84.85} & \textbf{77.18} & \textbf{80.37} & \textbf{81.65} & \textbf{84.34} & \textbf{71.11} & \textbf{74.40} & \textbf{78.21} & \textbf{81.28} & 80.63 & 83.50 & \makecell{\textbf{80.60}\\ \textcolor{red}{(+3.46)}} & \makecell{\textbf{83.35}\\ \textcolor{red}{(+3.15)}} \\
\midrule
\multicolumn{24}{c}{Satellite \ensuremath{\rightarrow} Drone}\\
\midrule
LRFR*~\cite{gan2025learning} & ConvNeXt & \textbf{95.72} & \textbf{93.22} & 87.02 & 59.76 & 83.74 & 43.81 & 77.03 & 45.11 & 53.78 & 16.47 & 48.50 & 10.62 & 80.31 & 39.72 & 57.35 & 15.42 & 89.02 & 69.65 & \textbf{93.15} & \textbf{83.29} & 76.56 & 47.71\\
Zheng \etal~\cite{zheng2020university} & ResNet-50 & 83.45 & 67.94 & 79.60 & 61.12 & 77.60 & 59.73 & 73.18 & 55.07 & 75.89 & 54.45 & 70.76 & 43.26 & 74.75 & 56.44 & 69.47 & 39.25 & 72.18 & 51.91 & 76.46 & 57.59 & 75.33 & 54.68\\
He \etal~\cite{he2016deep} & ResNet-101 & 85.73 & 71.79 & 82.45 & 66.46 & 81.46 & 65.68 & 79.74 & 61.72 & 79.74 & 60.59 & 74.75 & 50.31 & 80.17 & 62.61 & 75.32 & 45.37 & 79.60 & 58.21 & 82.31 & 64.67 & 80.13 & 60.74\\ 
Huang \etal~\cite{huang2017densely} & DenseNet121 & 83.74 & 70.34 & 82.31 & 66.32 & 81.17 & 65.23 & 78.60 & 60.33 & 79.46 & 61.66 & 74.61 & 51.14 & 78.46 & 61.68 & 74.47 & 47.88 & 74.32 & 55.26 & 78.32 & 61.63 & 78.55 & 60.15\\
Safe-Net*~\cite{lin2024self} & Vit-S & 91.22 & 86.06 & 90.04 & \textbf{85.43} & 71.12 & 68.56 & 73.26 & 45.62 & 68.23 & 41.78 & 49.32 & 34.72 & 61.07 & 29.86 & 73.15 & 43.08 & 88.54 & 74.65 & 90.02 & 78.21 & 75.69 & 58.80\\
Liu \etal~\cite{liu2021swin} & Swin-T & 80.74 & 68.94 & 81.03 & 67.46 & 81.17 & 66.39 & 78.46 & 61.33 & 79.17 & 64.65 & 74.89 & 56.57 & 78.89 & 63.49 & 75.61 & 48.43 & 76.60 & 56.57 & 78.74 & 64.45 & 78.53 & 61.83\\
Pan \etal~\cite{pan2018two} & IBN-Net & 86.31 & 73.54 & 84.59 & 67.61 & 84.74 & 69.03 & 80.88 & 64.44 & 83.31 & 63.71 & 77.89 & 52.14 & 83.02 & 65.74 & 78.46 & 50.77 & 79.46 & 58.64 & 84.02 & 67.94 & 82.27 & 63.36\\
LPN~\cite{wang2021each} & ResNet-50 & 87.02 & 75.19 & 86.16 & 71.34 & 83.88 & 69.49 & 82.88 & 65.39 & 84.59 & 66.28 & 79.60 & 55.19 & 84.17 & 66.26 & 82.88 & 52.05 & 81.03 & 62.24 & 84.14 & 67.35 & 83.64 & 65.08\\
Muse-Net~\cite{wang2024multiple} & ResNet-50 $\times 2$ & 88.02 & 75.10 & 87.87 & 69.85 & 87.73 & 71.12 & 83.74 & 66.52 & 85.02 & 67.78 & 80.88 & 54.26 & 84.88 & 67.75 & 80.74 & 53.01 & 81.60 & 62.09 & 86.31 & 70.03 & 84.68 & 65.75\\
\hline
Baseline~\cite{wen2025WeatherPrompt} & X-VLM & 89.16 & 81.80 & 88.73 & 80.58 & 88.16 & 79.87 & 87.59 & 77.25 & 88.45 & 78.20 & 86.73 & 73.23 & 88.59 & 78.14 & 86.59 & 65.20 & 85.31 & 73.25 & 87.88 & 76.33 & 87.72 & 76.39\\
Ours & X-VLM & 90.58 & 85.13 & \textbf{91.30} & 84.53 & \textbf{90.87} & \textbf{83.70} & \textbf{90.30} & \textbf{81.27} & \textbf{91.16} & \textbf{82.28} & \textbf{89.87} & \textbf{77.43} & \textbf{90.73} & \textbf{81.94} & \textbf{89.30} & \textbf{70.56} & \textbf{89.44} & \textbf{78.24} & 90.30 & 80.77 & \makecell{\textbf{90.39}\\ \textcolor{red}{(+2.67)}} & \makecell{\textbf{80.59}\\ \textcolor{red}{(+4.20)}} \\
\bottomrule
\end{tabular}
}
\end{table}

\begin{table}[!t]
\caption{Performance (R@1 ($\%$) and AP ($\%$)) on DenseUAV~\cite{dai2023vision} for Drone \ensuremath{\rightarrow} Satellite and Satellite \ensuremath{\rightarrow} Drone tasks under multiple weathers. Best results are in \textbf{bold}. \label{tab:table2}}
\vspace{-10pt}
\centering
\small
\renewcommand{\arraystretch}{1.0}  
\setlength{\tabcolsep}{1.2pt}
\resizebox{\linewidth}{!}{
\begin{tabular}{l|l|c c|c c|c c|c c|c c|c c|c c|c c|c c|c c|c c}
\toprule
\multirow{2}{*}{Method} & \multirow{2}{*}{Backbone} & \multicolumn{2}{c|}{Normal} & \multicolumn{2}{c|}{Fog} & \multicolumn{2}{c|}{Rain} & \multicolumn{2}{c|}{Snow} & \multicolumn{2}{c|}{Fog+Rain} & \multicolumn{2}{c|}{Fog+Snow} & \multicolumn{2}{c|}{Rain+Snow} & \multicolumn{2}{c|}{Dark} & \multicolumn{2}{c|}{Over-exp} & \multicolumn{2}{c|}{Wind} & \multicolumn{2}{c}{Mean}\\
& & R@1 & AP & R@1 & AP & R@1 & AP & R@1 & AP & R@1 & AP & R@1 & AP & R@1 & AP & R@1 & AP & R@1 & AP & R@1 & AP & R@1 & AP\\
\midrule
\multicolumn{24}{c}{Drone \ensuremath{\rightarrow} Satellite}\\
\midrule
Safe-Net~\cite{lin2024self} & Vit-S & 22.27 & 27.50 & 18.66 & 23.52 & 9.52 & 12.98 & 6.82 & 9.55 & 8.49 & 11.32 & 6.44 & 9.61 & 7.34 & 10.36 & 10.94 & 14.58 & 12.23 & 16.28 & 27.54 & 32.68 & 13.03 & 16.84\\
Muse-Net~\cite{wang2024multiple} & ResNet-50 $\times 2$ & 42.60 & 48.59 & 38.48 & 44.29 & 40.41 & 46.49 & 40.80 & 46.71 & 36.42 & 42.21 & 30.63 & 36.67 & 41.18 & 47.23 & 26.64 & 31.81 & 33.08 & 39.02 & 42.60 & 48.69 & 37.28 & 43.17\\
LRFR~\cite{gan2025learning} & ConvNeXt & 42.21 & 49.93 & 42.73 & 50.19 & 42.73 & 50.04 & 41.83 & 49.64 & 39.90 & 47.40 & 39.25 & 46.28 & 43.89 & 51.17 & 35.65 & 42.35 & 42.47 & 49.65 & 44.66 & 51.99 & 41.53 & 48.86\\
LPN~\cite{wang2021each} & ResNet-50 & 46.85 & 52.70 & 43.24 & 49.27 & 40.54 & 46.57 & 41.96 & 48.20 & 36.94 & 42.58 & 32.56 & 38.15 & 42.08 & 47.99 & 28.70 & 33.84 & 39.25 & 45.73 & 46.46 & 52.67 & 44.67 & 45.77\\
\hline
Baseline~\cite{wen2025WeatherPrompt} & X-VLM & 30.76 & 37.00 & 32.05 & 37.99 & 31.66 & 37.58 & 29.60 & 35.43 & 29.86 & 36.00 & 26.77 & 32.97 & 30.63 & 36.63 & 23.17 & 28.58 & 25.61 & 30.95 & 32.43 & 38.74 & 29.25 & 35.19\\
Ours & X-VLM & \textbf{54.57} & \textbf{60.72} & \textbf{53.80} & \textbf{59.89} & \textbf{54.95} & \textbf{60.62} & \textbf{54.31} & \textbf{60.03} & \textbf{53.41} & \textbf{59.25} & \textbf{50.06} & \textbf{56.18} & \textbf{54.18} & \textbf{60.18} & \textbf{46.07} & \textbf{51.47} & \textbf{47.36} & \textbf{53.90} & \textbf{55.60} & \textbf{61.70} &  \makecell{\textbf{52.43}\\ \textcolor{red}{(+23.18)}} & \makecell{\textbf{58.39}\\ \textcolor{red}{(+23.20)}} \\
\midrule
\multicolumn{24}{c}{Satellite \ensuremath{\rightarrow} Drone}\\
\midrule
Safe-Net~\cite{lin2024self} & Vit-S & 22.01 & 27.75 & 24.07 & 29.17 & 9.65 & 13.42 & 8.24 & 11.62 & 9.01 & 13.01 & 7.34 & 10.16 & 9.01 & 12.35 & 17.37 & 21.47 & 13.00 & 17.41 & 27.67 & 33.90 & 14.74 & 19.03\\
Muse-Net~\cite{wang2024multiple} & ResNet-50 $\times 2$ & 37.71 & 44.75 & 35.78 & 41.54 & 32.95 & 39.76 & 36.29 & 43.16 & 34.23 & 40.09 & 30.12 & 35.99 & 36.55 & 42.75 & 24.71 & 29.36 & 29.47 & 35.83 & 37.71 & 44.65 & 33.55 & 39.79\\
LRFR~\cite{gan2025learning} & ConvNeXt & 45.56 & 53.71 & 45.17 & 52.91 & 43.37 & 51.43 & 45.17 & 52.81 & 40.67 & 48.90 & 36.55 & 44.97 & 43.11 & 51.10 & 32.95 & 40.03 & 40.15 & 48.04 & 46.20 & 54.19 & 41.89 & 49.81\\
LPN~\cite{wang2021each} & ResNet-50 & 48.65 & 54.60 & 46.33 & 52.09 & 46.59 & 52.20 & 47.36 & 53.32 & 44.27 & 49.99 & 36.04 & 42.45 & 45.43 & 51.55 & 31.53 & 36.75 & 39.00 & 45.42 & 48.65 & 54.77 & 43.39 & 49.31\\
\hline
Baseline~\cite{wen2025WeatherPrompt} & X-VLM & 30.76 & 37.26 & 28.57 & 35.16 & 29.99 & 36.23 & 29.09 & 35.04 & 28.06 & 34.26 & 25.10 & 31.08 & 28.44 & 34.64 & 22.27 & 27.96 & 22.91 & 28.90 & 32.56 & 38.52 & 27.78 & 33.91\\
Ours & X-VLM & \textbf{53.02} & \textbf{58.53} & \textbf{49.29} & \textbf{55.52} & \textbf{52.38} & \textbf{57.92} & \textbf{50.19} & \textbf{55.82} & \textbf{48.65} & \textbf{54.82} & \textbf{46.46} & \textbf{52.86} & \textbf{49.81} & \textbf{55.90} & \textbf{42.34} & \textbf{48.17} & \textbf{45.30} & \textbf{51.37} & \textbf{52.90} & \textbf{58.64} & \makecell{\textbf{49.03}\\ \textcolor{red}{(+21.25)}} & \makecell{\textbf{54.96}\\ \textcolor{red}{(+21.05)}} \\
\bottomrule
\end{tabular}
}
\vspace{-14pt}
\end{table}

\subsection{Comparison with Competitive Methods}
\noindent\textbf{Quantitative Results.}  Experimental results on the non-dense dataset University-1652 are summarized in~\cref{tab:table1}. We compare the proposed GeoFuse with several representative and competitive drone-based geo-localization methods, where the drone images are augmented with 10 distinct weather style transformations while satellite images remain unaltered. In the Drone \ensuremath{\rightarrow} Satellite retrieval task, GeoFuse achieves a mean accuracy of R@1 of $80.60\%$ and a mean AP of $83.35\%$, surpassing the previous state-of-the-art methods by $3.46\%$ and $3.15\%$, respectively. This improvement validates the superiority of our geometric-guided fusion design, which effectively enhances the discriminability of drone features under multi-weather conditions. Similarly, in the Satellite \ensuremath{\rightarrow} Drone retrieval task, our method attains a mean accuracy of R@1 of $90.39\%$ and a mean AP of $80.59\%$, outperforming existing approaches by $2.67\%$ and $4.20\%$, respectively. These consistent gains across both retrieval directions confirm that GeoFuse can robustly align drone and satellite representations even when large visual variations are introduced by complex weather conditions.

We observe a similar result on the dense dataset DenseUAV in~\cref{tab:table2}. All comparative methods were retrained on DenseUAV with their official implementations from GitHub to ensure a fair comparison, with both training and testing conducted on images captured at an altitude of 80 m. For the Drone \ensuremath{\rightarrow} Satellite retrieval scenario, GeoFuse outperforms all competitive methods, boosting the mean R@1 accuracy from $29.25\%$ to $52.43\%$ $(+23.18\%)$ and the mean AP from $35.19\%$ to $58.39\%$ $(+23.20\%)$. For the Satellite \ensuremath{\rightarrow} Drone retrieval scenario, the proposed method achieves similarly superior performance, increasing the mean R@1 accuracy from $27.78\%$ to $49.03\%$ and the mean AP from $33.91\%$ to $54.96\%$, yielding significant gains of $+21.25\%$ in mean R@1 and $+21.05\%$ in mean AP. These results validate excellent generalization of GeoFuse across datasets with different spatial sampling densities. The results show that the compared methods exhibit highly unstable performance on this dense sampling dataset, which differs significantly from their performance on University-1652.  Specifically, the baseline method WeatherPrompt, which performed excellently on University-1652, suffers a substantial performance drop on DenseUAV, falling below many methods that were inferior to it on University-1652. 
Conversely, several methods that underperformed on University-1652 show relative performance enhancements on this dataset, reflecting the large performance fluctuations of existing methods across datasets with different characteristics. In contrast, GeoFuse consistently attains competitive results, demonstrating its robustness and stable generalization across datasets.

\noindent\textbf{Discussion. Difference between non-dense and dense datasets.} 
The performance instability mainly stems from the dense sampling in dense datasets, where adjacent viewpoints share highly similar appearances due to small spatial intervals. Most existing methods, like ours, heavily rely on fine-grained local visual details for discriminative feature learning. However, these methods are overly sensitive to local noise (\eg, rain, snow, or other weather variations), causing severe degradation in dense scenarios where fine differences become unreliable. In contrast, GeoFuse achieves greater robustness by fusing satellite imagery with road maps. While still utilizing fine-grained features from satellite/drone views, the road map introduces accurate, stable geographic structural constraints that are largely unaffected by visual noise. This complementary global and structural guidance helps the model learn geographically discriminative representations, maintaining strong performance across both non-dense (\eg, University-1652) and dense datasets without being misled by local appearance changes.

\begin{figure}[!t]
\centering
\vspace{-.2in}
\includegraphics[width=1\textwidth]{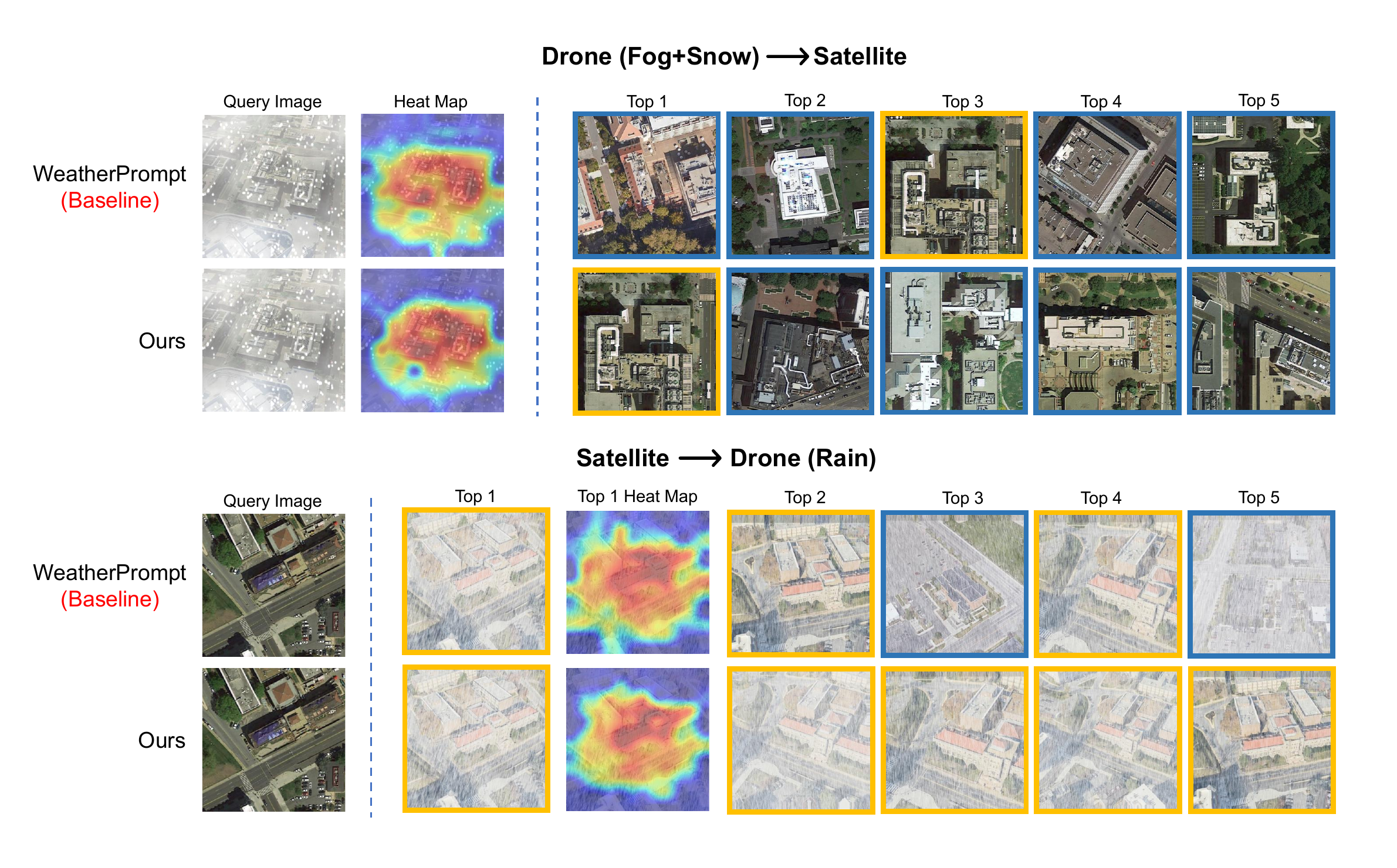}
\vspace{-.4in}
\caption{Qualitative comparison on University-1652. We show heatmaps and top-5 retrieval results between our method and the baseline under different weather conditions. The correct matches are highlighted in yellow boxes, while the incorrect matches are enclosed in blue boxes.}
\vspace{-.2in}
\label{fig_4}
\end{figure}

\noindent\textbf{Qualitative Results.}  
We further conduct a qualitative comparison with the baseline method, including heatmap visualizations and top-5 retrieval results (see~\cref{fig_4}). 
Thanks to the learned weather-invariant representations, our method produces heatmaps with finer-grained and more robust activation patterns in adverse weather conditions. Specifically, the attention is effectively concentrated on distinctive structural elements (\eg, buildings), while avoiding spurious focus on transient noise such as raindrops or snowflakes. As a result, the activation remains sharply localized on semantically relevant regions, with significantly less dispersion toward weather-induced artifacts compared to the baseline. In contrast, the baseline method often generates more diffused activation maps, with noticeable attention leaking to noise patterns in challenging weather, which weakens its discriminative power. This focused yet robust attention pattern in our approach is consistently observed in both drone-to-satellite and satellite-to-drone retrieval directions.


\subsection{Ablation Studies and Further Discussion}
\noindent\textbf{Effect of Cross-modal Fusion and Class-level Contrastive Learning.} To analyze the contribution of each proposed component, we perform an ablation study on University-1652. Specifically, we evaluate the effects of (1) token-level fusion, (2) channel-level fusion, and (3) class-level contrastive loss, individually and in combination.~\cref{tab:table3} summarizes the results in terms of mean R@1 and mean AP. We observe that adding each component progressively boosts performance. Token-level fusion establishes a strong baseline, while the addition of channel-level fusion and the class-level contrastive loss further enhances the discriminability of the learned representations. Overall, the full model combining token-level fusion, channel-level fusion, and class-level contrastive loss achieves gains of $+0.73\%$ in R@1 accuracy and $+0.71\%$ in mean AP compared to employing token-level fusion only.

\noindent\textbf{Effect of Class-level Contrastive Loss Hyper-parameters.} We further investigate the effect of the hyper-parameter $\lambda$, which controls the contribution of the class-level cross-view contrastive loss in the overall training objective. An ablation study is conducted on University-1652 with $\lambda$ values set to $0.05$, $0.10$, and $0.15$, as summarized in~\cref{tab:table4a}. The results show that $\lambda = 0.10$ yields the optimal performance in terms of R@1 and AP, indicating an optimal balance between the contribution of the class-level contrastive loss and the other components of the training objective. Setting $\lambda$ too low weakens the supervision effect of the contrastive loss, while a larger value overemphasizes it, which may adversely affect the learning of other objectives. This analysis validates the benefit of properly weighting the class-level contrastive loss for effective cross-view feature alignment. We also evaluate the temperature $\tau$ with values $0.05$, $0.07$, and $0.10$ (see~\cref{tab:table4b}). The default setting $\tau = 0.07$ yields the highest performance, while the results with $\tau = 0.05$ and $0.10$ remain comparable. This indicates that our method is relatively insensitive to the choice of the temperature hyper-parameter and exhibits promising scalability across different settings.

\begin{table}[!t]
\caption{Ablation study on University-1652 about token-level fusion combined with channel-level fusion and class-level contrastive loss, respectively. \label{tab:table3}}
\vspace{-10pt}
\centering
\fontsize{6pt}{8pt}\selectfont
\setlength{\tabcolsep}{3pt} 
\renewcommand{\arraystretch}{0.85}
\resizebox{\linewidth}{!}{
\begin{tabular}{l cc c cc} 
\noalign{\hrule height 0.6pt}
& \multicolumn{2}{c}{D2S} & & \multicolumn{2}{c}{S2D} \\ 
\cline{2-3} \cline{5-6} 
Method & Mean R@1 ($\%$) & Mean AP ($\%$) & & Mean R@1 ($\%$) & Mean AP ($\%$) \\
\hline 
Token-level Fusion Only & 79.87 & 82.64 & & 89.90 & 79.25 \\ 
$+$ Channel-level Fusion & 80.11 & 82.84 & & 89.74 & 80.12 \\ 
$+$ Class-level Contrastive Loss & \textbf{80.60} & \textbf{83.35} & & \textbf{90.39} & \textbf{80.59} \\ 
\noalign{\hrule height 0.6pt}
\end{tabular}
}
\vspace{-.15in}
\end{table}

\begin{table}[!t]
\centering
\caption{Ablation of the hyper-parameters in the class-level contrastive loss on University-1652. \label{tab:table4}}
\vspace{-10pt}
\begin{subtable}[t]{0.48\linewidth}
\centering
\resizebox{\linewidth}{!}{
\begin{tabular}{l cc c cc} 
\toprule
& \multicolumn{2}{c}{D2S} & & \multicolumn{2}{c}{S2D} \\
\cline{2-3} \cline{5-6} 
$\lambda$ & Mean R@1 ($\%$) & Mean AP ($\%$) & & Mean R@1 ($\%$) & Mean AP ($\%$) \\
\hline
0.05 & 79.96 & 82.81 & & \textbf{90.67} & 79.94 \\
0.10 & \textbf{80.60} & \textbf{83.35} & & 90.39 & \textbf{80.59} \\
0.15 & 79.94 & 82.75 & & 89.74 & 79.51 \\
\bottomrule
\end{tabular}
}
\caption{Effect of $\lambda$} 
\label{tab:table4a}
\end{subtable}
\hfill 
\begin{subtable}[t]{0.48\linewidth}
\centering
\resizebox{\linewidth}{!}{
\begin{tabular}{l cc c cc} 
\toprule
& \multicolumn{2}{c}{D2S} & & \multicolumn{2}{c}{S2D} \\
\cline{2-3} \cline{5-6} 
$\tau$ & Mean R@1 ($\%$) & Mean AP ($\%$) & & Mean R@1 ($\%$) & Mean AP ($\%$) \\
\hline
0.05 & 79.09 & 82.06 & & 89.08 & 79.29 \\
0.07 & \textbf{80.60} & \textbf{83.35} & & \textbf{90.39} & \textbf{80.59} \\
0.10 & 79.78 & 82.64 & & 90.33 & 79.89 \\
\bottomrule
\end{tabular}
}
\caption{Effect of $\tau$}
\label{tab:table4b}
\end{subtable}
\fontsize{6pt}{8pt}\selectfont
\setlength{\tabcolsep}{3pt} 
\renewcommand{\arraystretch}{0.85}
\vspace{-0.4in}
\end{table}

\noindent\textbf{Effect of Geometric-guided Auxiliary Modalities.} 
To investigate the impact of different geometric-guided auxiliary modalities fused with satellite images, we conduct an ablation study by comparing three variants: (1) road maps containing textual annotations, (2) road maps with all text removed, and (3) pseudo-depth maps generated from satellite images by Depth Anything~\cite{yang2024depth}. As summarized in~\cref{tab:table5}, the pseudo depth maps achieve the lowest performance in both mean R@1 and AP. Since they are directly derived from satellite images, pseudo-depth maps carry highly restricted and redundant information, which cannot provide richer geometric cues compared with road maps. The road maps without text show slightly lower overall performance than the ones with text. Nevertheless, they achieve the highest AP in the Satellite \ensuremath{\rightarrow} Drone task, while avoiding potential location label leakage from textual annotations. These results validate that the choice of auxiliary modality significantly affects performance and that the constructed text-free road maps in this work offer a favorable trade-off between boosting localization accuracy and mitigating location leakage from textual annotations.

\begin{table}[!t]
\caption{Ablation study of different geometric-guided auxiliary modalities on University-1652. \label{tab:table5}}
\vspace{-10pt}
\centering
\small
\setlength{\tabcolsep}{2pt} 
\resizebox{\linewidth}{!}{
\begin{tabular}{l|c c|c c|c c|c c|c c|c c|c c|c c|c c|c c|c c}
\toprule
\multirow{2}{*}{Modality} & \multicolumn{2}{c|}{Normal} & \multicolumn{2}{c|}{Fog} & \multicolumn{2}{c|}{Rain} & \multicolumn{2}{c|}{Snow} & \multicolumn{2}{c|}{Fog+Rain} & \multicolumn{2}{c|}{Fog+Snow} & \multicolumn{2}{c|}{Rain+Snow} & \multicolumn{2}{c|}{Dark} & \multicolumn{2}{c|}{Over-exp} & \multicolumn{2}{c|}{Wind} & \multicolumn{2}{c}{Mean}\\
& R@1 & AP & R@1 & AP & R@1 & AP & R@1 & AP & R@1 & AP & R@1 & AP & R@1 & AP & R@1 & AP & R@1 & AP & R@1 & AP & R@1 & AP\\
\midrule
\multicolumn{23}{c}{Drone \ensuremath{\rightarrow} Satellite}\\
\midrule
Pseudo Depth Maps & 85.68 & 87.72 & 84.48 & 86.67 & 83.56 & 85.92 & 80.48 & 83.23 & 81.98 & 84.54 & 76.17 & 79.40 & 81.42 & 84.05 & 71.00 & 74.56 & 77.70 & 80.78 & 80.54 & 83.30 & 80.30 & 83.02\\
Road Maps $w$ Text & \textbf{85.97} & \textbf{88.09} & 84.66 & 86.95 & \textbf{84.12} & \textbf{86.49} & 81.23 & 84.03 & \textbf{82.31} & \textbf{84.92} & 77.09 & 80.32 & \textbf{82.12} & \textbf{84.77} & 71.06 & \textbf{74.63} & 77.85 & 81.05 & \textbf{81.18} & \textbf{84.00} & \textbf{80.76} & \textbf{83.53}\\
Road Maps $w/o$ Text & 85.26 & 87.44 & \textbf{84.72} & \textbf{86.96} & 83.61 & 86.02 & \textbf{81.38} & \textbf{84.08} & 82.26 & 84.85 & \textbf{77.18} & \textbf{80.37} & 81.65 & 84.34 & \textbf{71.11} & 74.40 & \textbf{78.21} & \textbf{81.28} & 80.63 & 83.50 & 80.60 & 83.35\\
\midrule
\multicolumn{23}{c}{Satellite \ensuremath{\rightarrow} Drone}\\
\midrule
Pseudo Depth Maps & 90.01 & 85.06 & \textbf{91.73} & 83.64 & 90.01 & 83.03 & 89.02 & 80.00 & 89.87 & 81.20 & 89.44 & 75.51 & 90.01 & 80.93 & 87.02 & 69.05 & 86.59 & 76.16 & 89.16 & 79.88 & 89.29 & 79.45\\
Road Maps $w$ Text & \textbf{91.87} & \textbf{85.50} & 91.01 & 84.47 & \textbf{91.16} & \textbf{83.85} & \textbf{90.30} & 81.09 & 91.01 & 82.21 & 89.16 & 76.81 & \textbf{90.73} & 81.82 & \textbf{89.44} & 70.11 & 89.30 & 77.46 & \textbf{90.44} & \textbf{80.84} & \textbf{90.44} & 80.42\\
Road Maps $w/o$ Text & 90.58 & 85.13 & 91.30 & \textbf{84.53} & 90.87 & 83.70 & \textbf{90.30} & \textbf{81.27} & \textbf{91.16} & \textbf{82.28} & \textbf{89.87} & \textbf{77.43} & \textbf{90.73} & \textbf{81.94} & 89.30 & \textbf{70.56} & \textbf{89.44} & \textbf{78.24} & 90.30 & 80.77 & 90.39 & \textbf{80.59}\\
\bottomrule
\end{tabular}
}
\vspace{-.25in}
\end{table}

\noindent\textbf{Limitations.} Despite the superior performance, our method still has certain limitations that warrant future investigation. It relies on the availability and precise spatial alignment of road maps with satellite imagery, which can be unavailable or misaligned in remote mountainous regions, newly developed urban areas, conflict zones, or scenes affected by recent infrastructure changes and low-resolution data, thereby impairing cross-modal fusion. More critically, in extremely unstructured environments (e.g., deserts, forests) where road networks are sparse or entirely absent, the geometric guidance degrades to effectively blank input, resulting in reduced feature discriminability and degraded geo-localization accuracy. Nevertheless, our carefully designed training process, incorporating strong regularization, augmentation, and loss formulations, provides effective safeguards against overfitting and ensures reasonable model stability even under such complete guidance failure, preventing catastrophic performance collapse. 

\section{Conclusion}
In this work, we present GeoFuse, a robust drone-view geo-localization framework that exploits freely available road maps as a weather-invariant geometric prior to counter adverse weather effects. By fusing precisely aligned road-map and satellite features via token- and channel-level interactions, our adaptive fusion module effectively balances modality contributions, while class-level cross-view contrastive learning aligns degraded drone views with the fused representations. Experiments on University-1652 and DenseUAV verify consistent gains in localization accuracy across diverse weather conditions and strong cross-dataset generalization. These results highlight the value of road maps as lightweight, accessible auxiliary priors, opening a promising path toward reliable multi-weather drone geo-localization.



%
%
\bibliographystyle{splncs04}
\bibliography{main}
\end{document}